\documentclass[conference,letterpaper]{IEEEtran}
\usepackage{cite}
\usepackage{amsmath,amssymb,amsfonts}
\usepackage{algorithmic}
\usepackage{graphicx}
\usepackage{textcomp}
\usepackage{pgfplots}
\usepackage{booktabs}
\usepackage{balance}
\usepackage{multicol}
\usepackage{float}
\usepackage{tabularx}
{\usepackage{multirow}
\usepackage{color, colortbl} 

\begin{document}
\title{Demo: A Practical Testbed for Decentralized Federated Learning on Physical Edge Devices}

\author{
    \IEEEauthorblockN{Chao Feng\IEEEauthorrefmark{1}, Nicolas Huber\IEEEauthorrefmark{1}, Alberto Huertas Celdrán\IEEEauthorrefmark{1}, G\'er\^ome Bovet\IEEEauthorrefmark{2}, Burkhard Stiller\IEEEauthorrefmark{1}}
    
    \IEEEauthorblockA{\IEEEauthorrefmark{1}Communication Systems Group, Department of Informatics, University of Zurich UZH, CH--8050 Zürich, Switzerland \\{[cfeng, huertas, stiller]}@ifi.uzh.ch, nicolas.huber3@uzh.ch}
    \IEEEauthorblockA{\IEEEauthorrefmark{2}Cyber-Defence Campus, armasuisse Science \& Technology, CH--3602 Thun, Switzerland gerome.bovet@armasuisse.ch}
}

\maketitle

\begin{abstract}
Federated Learning (FL) enables collaborative model training without sharing raw data, preserving participant privacy. Decentralized FL (DFL) eliminates reliance on a central server, mitigating the single point of failure inherent in the traditional FL paradigm, while introducing deployment challenges on resource-constrained devices. To evaluate real-world applicability, this work designs and deploys a physical testbed using edge devices such as Raspberry Pi and Jetson Nano. The testbed is built upon a DFL training platform, NEBULA, and extends it with a power monitoring module to measure energy consumption during training. Experiments across multiple datasets show that model performance is influenced by the communication topology, with denser topologies leading to better outcomes in DFL settings.
\end{abstract}


\IEEEpeerreviewmaketitle

\section{Introduction}
With the proliferation of advanced sensors, enhanced processing capabilities, and widespread connectivity, edge devices now generate vast amounts of data, offering significant opportunities for Machine Learning (ML) \cite{DBLP:journals/corr/abs-2002-10610}. However, the distributed nature of data from IoT devices makes traditional centralized ML approaches impractical due to privacy concerns, communication overhead, and regulatory constraints \cite{Mart_nez_Beltr_n_2023}. Federated Learning (FL), first introduced by \cite{konevcny2016federated}, addresses these issues by enabling devices to collaboratively train a shared model without sharing raw data, exchanging only model updates instead. FL can be categorized into Centralized Federated Learning (CFL), where a central server coordinates training and aggregation, and Decentralized Federated Learning (DFL), where nodes interact directly without a central authority. While CFL is widely used, DFL offers better scalability, fault tolerance, and eliminates single points of failure and trust bottlenecks \cite{Mart_nez_Beltr_n_2023}.

Training FL models on resource-constrained devices, such as smartphones, IoT devices, and edge computing nodes, introduces significant challenges due to their limited computational power, memory, and energy capacities, complicating the implementation of privacy-preserving methods and efficient model training \cite{9916265}. Current DFL platforms, such as NEBULA ~\cite{beltran2024fedstellar}, primarily rely on simulation-based or virtualization techniques, like Docker containers, for development and testing. However, these approaches fail to adequately reflect the constraints and complexities encountered when deploying FL on heterogeneous, resource-constrained hardware, using diverse datasets and varying overlay networks. 

This work addresses this research gap by designing and implementing a realistic, physical DFL testbed utilizing heterogeneous resource-constrained devices, including multiple models of Raspberry Pi and NVIDIA developer kits such as the Jetson series, to replicate real-world deployment scenarios closely. The testbed supports experimentation across multiple topological configurations, enabling comprehensive evaluations with diverse datasets. Compared with simulation-based or virtualization-based platforms, the proposed physical testbed achieves comparable model training performance while accurately reflecting device heterogeneity, hardware constraints, and network topologies, thus providing a robust and realistic environment for evaluating decentralized federated learning frameworks. Additionally, the designed testbed incorporates mechanisms to record energy consumption throughout the model training process, laying a foundation for subsequent research on DFL systems' sustainability and energy efficiency.

\section{Testbed Design}
This section details the construction of a physical testbed designed to evaluate DFL under realistic hardware and network conditions. Rather than relying on simulated or virtualized setups, the testbed is built upon actual, heterogeneous edge devices with constrained resources. It incorporates Raspberry Pi 4 Model B units (two with 4GB RAM and one with 2GB RAM, running Debian GNU/Linux 12) and an NVIDIA Jetson Nano Developer Kit (running Ubuntu 18.04.6 LTS), interconnected via a local Ethernet network.

\begin{figure}[t]
\centering
\includegraphics[width=0.45\textwidth]{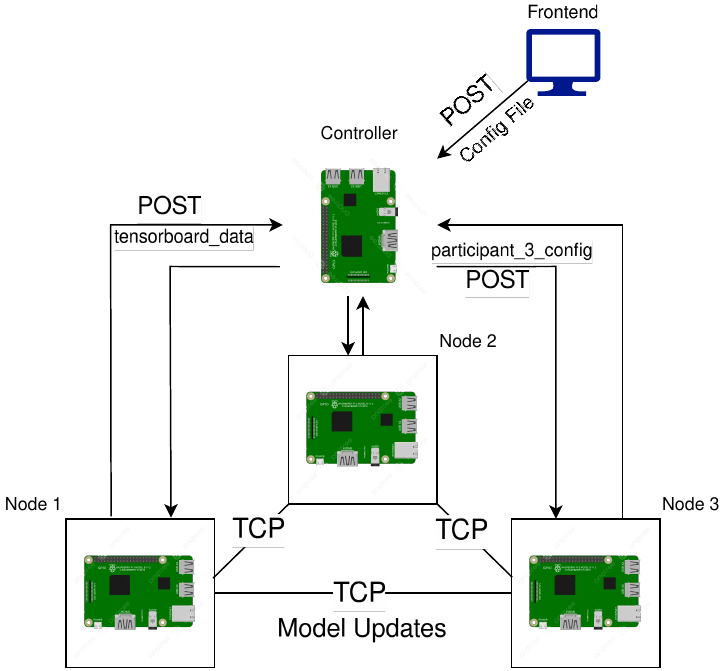}
\caption{System architecture showing FL configuration and metric transmission.}
\label{fig:architecture}
\end{figure}

\begin{figure}[t]
\centering
\includegraphics[width=0.45\textwidth]{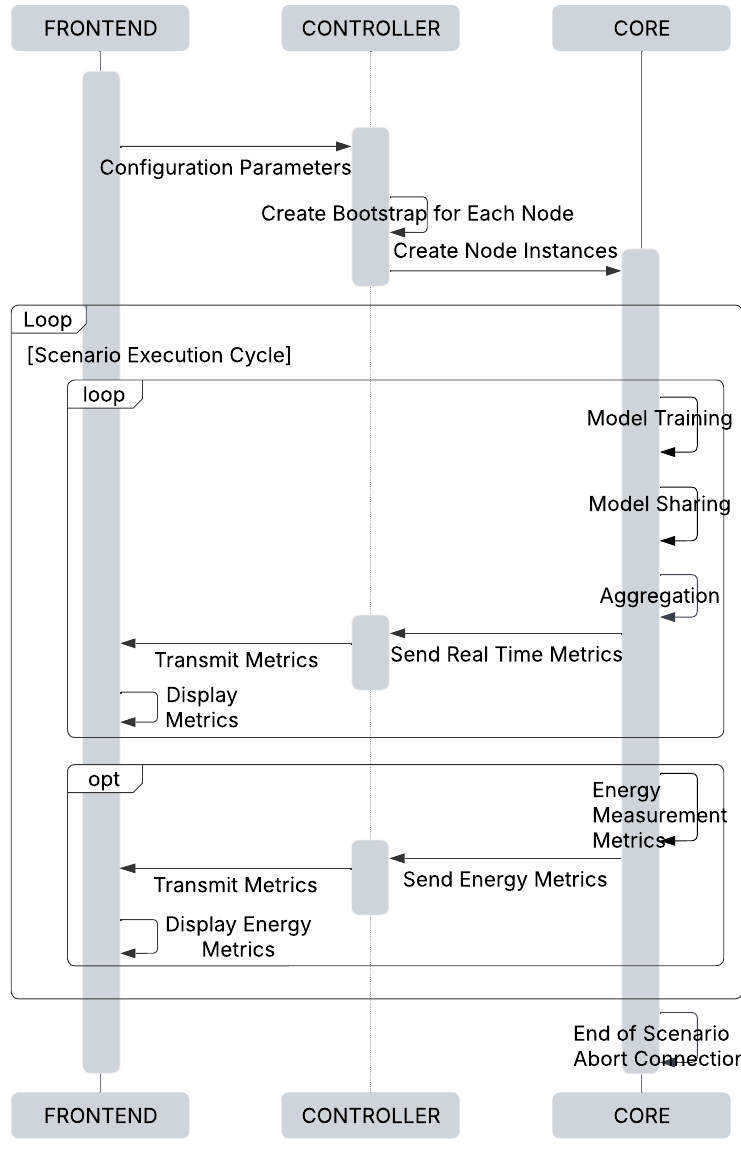}
\caption{Sequence diagram of configuration distribution, model training, and metric reporting.}
\label{fig:sequence_diag}
\end{figure}

A DFL platform, \textit{i.e.}, NEBULA, serves as the underlying infrastructure for model training, aggregation, and inter-device communication within the testbed. Originally, NEBULA employs Docker containers to emulate DFL nodes, necessitating several modifications for physical hardware deployment:

\begin{itemize}
\item Deploying lightweight HTTP servers on each device to securely receive FL configurations from a controller.
\item Implementing periodic HTTP-based transmission of real-time metrics (e.g., CPU, memory usage) from nodes to the controller, facilitating continuous frontend visualization.
\item Extending monitoring capabilities to track energy consumption using JT-TC66C USB multimeters.
\end{itemize}

Figure~\ref{fig:architecture} depicts the testbed architecture. The controller receives configuration files from the frontend and distributes them to participant nodes. Each node hosts a local HTTP server to accept the configuration, followed by the server's termination to release the port. Nodes then establish TCP connections with their neighbors and begin decentralized training.

\begin{table}[b]
\centering
\caption{Comparison of validation scenarios.}
\label{tab:validation_scenario}
\resizebox{0.45\textwidth}{!}{%
    \begin{tabular}{@{} lll @{}}
    \toprule
    \textbf{Characteristic} & \textbf{Physical Scenario} & \textbf{Virtualized Scenario} \\
    \hline
    \multirow{2}{*}{\textbf{Participant}} & 3 Raspberry Pi 4,  & 4 Docker containers \\

    & 1 NVIDIA Jetson Nano & \\
    \hline
    \multirow{2}{*}{\textbf{Dataset}} & MNIST & MNIST \\
    & FashionMNIST & FashionMNIST \\
    \hline
    \multirow{2}{*}{\textbf{Topology}} & Fully connected, & Fully Connected \\
    & Star, Ring, Random \\
    \bottomrule
    \end{tabular}%
}
\end{table}

Figure~\ref{fig:sequence_diag} presents the scenario execution sequence: the frontend initiates the experiment by sending configuration parameters to the controller, which bootstraps node instances and distributes configuration files. During the training process, nodes periodically transmit training metrics to the controller, which are then visualized on the frontend. Additionally, if energy monitoring is enabled, nodes log power consumption throughout the training cycle and transmit a summary at the end of the scenario. The JT-TC66C multimeters measure power consumption in real time and provide both periodic metric updates and post-training batch reports. This dual-reporting mechanism enables accurate sustainability assessments without interrupting model training.
\section{Setup and Demonstration}
To evaluate the performance and feasibility of the proposed DFL testbed, a series of experiments were conducted under two deployment scenarios: Physical Scenario and Virtualized Scenario.

\begin{table*}[t]
\centering
\caption{Summary of Experimental Results for Physical and Virtualized Scenarios}
\label{tab:results_summary}
\begin{tabular}{l|c|c|c|c|c|c}
\hline
\textbf{Scenario} & \textbf{Avg. F$_{1}$-Score} & \textbf{CPU Usage (\%)} & \textbf{RAM Usage (\%)} & \textbf{Net Traffic (MB)} & \textbf{Power (W)} & \textbf{Energy (J)} \\
\hline
Fully Conn. MNIST (Phys.) & 82.0 & 25.7 & 33.3 & 30 & 3.3 & 1404 \\
Fully Conn. Fashion (Phys.) & 81.0 & 28.7 & 33.3 & 30 & 3.5 & 1350 \\
Star MNIST (Phys.) & 76.0 & 25.8 & 32.8 & P0: 30, Oth: 10 & 3.0 & 1251 \\
Star Fashion (Phys.) & 75.0 & 29.1 & 33.9 & P0: 30, Oth: 10 & 3.5 & 1324 \\
Ring MNIST (Phys.) & 78.5 & 26.0 & 32.9 & 20 & 3.05 & 1249 \\
Ring Fashion (Phys.) & 68.8 & 28.3 & 33.1 & 20 & 3.45 & 1404 \\
Random MNIST (Phys.) & 74.7 & 26.1 & 33.4 & Varied & 3.1 & 1305 \\
Random Fashion (Phys.) & 64.9 & 28.7 & 33.0 & Varied & 3.5 & 1368 \\ \hline
Fully Conn. MNIST (Virt.) & 82.0 & 26.8 & 7.4 & 30 & N/A & N/A \\
Fully Conn. Fashion (Virt.) & 81.0 & 34.4 & 7.4 & 30 & N/A & N/A \\
\hline
\end{tabular}
\end{table*}

\subsection{Experiments Setup}

Table~\ref{tab:validation_scenario} provides an overview of the experimental configuration in both settings. The physical testbed includes four edge devices: three Raspberry Pi 4 Model B boards and one NVIDIA Jetson Nano. This configuration introduces heterogeneity in terms of CPU and GPU capabilities. The Jetson Nano features a 128-core Maxwell GPU and a quad-core Cortex-A57 CPU (1.43GHz), whereas the Raspberry Pi 4 uses a Cortex-A72 CPU (1.5GHz). In contrast, the virtualized setup uses Docker containers deployed on a machine with an AMD Ryzen 7 5825U processor, 16GB RAM, and 512GB SSD, ensuring consistent and stable performance.

Both scenarios use MNIST \cite{6296535} and FashionMNIST \cite{xiao2017/online}, two widely recognized benchmarks for image classification. A simple multi-layer perceptron (MLP) model was used across all experiments, and the FedAvg \cite{konevcny2016federated} algorithm was trained over 10 federation rounds with one local epoch per round. To simulate realistic data heterogeneity, all datasets were partitioned independently and identically distributed (IID).

To assess the testbed’s flexibility and DFL’s sensitivity to communication structures, the physical setup tested four topologies: fully connected, star, ring, and random. Each topology was evaluated using both datasets. The virtualized baseline used only the fully connected topology due to simplicity and consistency constraints.

\subsection{Experimental Results}
The evaluation metrics include model F$_{1}$-score, CPU and RAM usage, network traffic, power draw, and energy consumption. Results from all physical experiments were compared against a virtualized baseline to assess both performance and resource efficiency.

As shown in Table~\ref{tab:results_summary}, the fully connected topology consistently achieved the highest average F$_{1}$-scores (81--82\%) on both MNIST and Fashion-MNIST datasets. In contrast, sparser topologies like ring and random exhibited more variability and generally lower average performance. The star topology showed that central nodes benefited from higher communication loads, often yielding higher accuracy than edge nodes.

Resource consumption reflected the underlying training cycles. Average CPU usage ranged from 25\% to 29\%, with peak usage exceeding 40\%. Power consumption followed similar patterns, ranging from 3 to 3.5 watts, and total energy use varied between 1251 and 1404 joules. RAM utilization remained consistent across experiments, with one Raspberry Pi (2 GB) showing higher usage due to memory constraints. Bandwidth usage aligned with topological structure: fully connected and star topologies generated higher traffic per node, while ring and random topologies maintained lower and more localized communication overhead.

The virtualized scenario matched physical F$_{1}$-scores but completed training significantly faster, highlighting the computational limits of edge devices. Despite lower training speeds, the physical testbed effectively replicated learning behavior, validating its feasibility for DFL research under realistic conditions.

\section{Conclusion}
This work presents a realistic, energy-aware testbed for evaluating DFL on heterogeneous, resource-constrained edge devices. Built on top of the NEBULA platform, the testbed integrates REST-based communication and real-time monitoring, including power and energy tracking using JT-TC66C multimeters. Experiments conducted across multiple topologies and two benchmark datasets (MNIST and Fashion-MNIST) demonstrate that accurate models can be trained on physical devices with limited computational resources, achieving performance comparable to virtualized environments. Resource usage, CPU, memory, bandwidth, and energy, remained within practical limits, confirming the testbed's suitability for realistic DFL research. Future extensions include scaling to larger federations, exploring more advanced topologies, optimizing training efficiency, and testing deployment in real-world IoT and mobile scenarios.

\ifCLASSOPTIONcompsoc
  \section*{Acknowledgments}
\else
  \section*{Acknowledgment}
\fi

This work has been partially supported by \textit{(a)} the Swiss Federal Office for Defense Procurement (armasuisse) with the CyberDFL project (CYD-C-2020003) and \textit{(b)} the University of Zürich, UZH.

\bibliographystyle{IEEEtran}
\balance
\bibliography{reference}

\end{document}